\documentclass{article}
\usepackage{spconf,amsmath,graphicx}
\usepackage{times}
\usepackage{latexsym}
\usepackage{graphicx} 
\usepackage{amsmath}
\usepackage{amsfonts}
\usepackage{bm}
\usepackage[linesnumbered,ruled,vlined]{algorithm2e}
\usepackage{comment}
\usepackage{microtype}
\usepackage{url}

\usepackage{hyperref}
\usepackage{arydshln}
\usepackage{multicol,multirow}

\title{Enhancing Unsupervised Speech Recognition with Diffusion GANs\thanks{Accepted by \href{https://2023.ieeeicassp.org/}{ICASSP 2023}}}
\name{Xianchao Wu} 
\address{NVIDIA \\ \href{xianchaow@nvidia.com}{xianchaow@nvidia.com}, \href{wuxianchao@gmail.com}{wuxianchao@gmail.com}}
\begin{document}
%
\maketitle
\begin{abstract}
We enhance the vanilla adversarial training method for unsupervised Automatic Speech Recognition (ASR) by a diffusion-GAN. Our model (1) injects instance noises of various intensities to the generator's output and unlabeled reference text which are sampled from pretrained phoneme language models with a length constraint, (2) asks diffusion timestep-dependent discriminators to separate them, and (3) back-propagates the gradients to update the generator. Word/phoneme error rate comparisons with wav2vec-U under Librispeech (3.1\% for test-clean and 5.6\% for test-other), TIMIT and MLS datasets, show that our enhancement strategies work effectively. 
\end{abstract}
\section{Introduction and Background}
\label{sec:intro}


\subsection{Unsupervised ASR}
\label{subsec:uasr}

Our work follows wav2vec-U \cite{wav2vecu_simp_NEURIPS2021_ea159dc9}, short for wav2vec Unsupervised, which utilizes wav2vec's self-supervised speech representations \cite{NEURIPS2020_92d1e1eb_wav2vec2_simp} to segment unlabeled audio and learns a mapping from these representations to phonemes via adversarial training. There are two lines (Figure \ref{fig:diffgan_wav2vecU_framework}) in the wav2vec-U framework. The first line accepts unlabeled speech audios as input, asks wav2vec2.0 \cite{NEURIPS2020_92d1e1eb_wav2vec2_simp} to encode audios into speech representations, and concurrently performs k-means to cluster representations of each time-step for audio segmentation and compresses the representations by Principal Component Analysis (PCA). Then, the spans of segments with similar cluster ids are merged together by mean pooling. A second mean pooling operation is performed between adjacent segments to balance the length aligning with future phoneme sequences. A result segment sequence $S$ is sent to a generator ($\mathcal{G}$, a single-layer non-causal convolutional network) to ``predict'' a phoneme sequence $\mathcal{G}(S)$. The second line first phonemizes unlabeled text into a ``pseudo'' reference phoneme sequence $P^r$ and then reshapes it into a one-hot representation. These two lines flow together to a discriminator ($\mathcal{C}$, a three-layer convolutional network) to perform adversarial training.

In wav2vec-U's objective setup, $\mathcal{G}$ and $\mathcal{C}$ are alternatively optimized with different batches of $(S, P^r)$ pairs. In the generator $\mathcal{G}$'s turn, the loss function to be minimized is:
\begin{equation}
    \mathcal{L}_\mathcal{G}=\text{BCE}(\mathcal{C}(\mathcal{G}(S)), 0) + \eta\mathcal{L}_{pd} + \gamma\mathcal{L}_{sp}. \label{eq:loss_g}
\end{equation}
Here, BCE stands for binary cross entropy loss. $\mathcal{L}_{pd}$ is a phoneme diversity loss which tries to balance the usage of the phoneme set through penalizing $\mathcal{G}$'s low usage of the phoneme entries on the batch $B$ level. The entropy of the averaged softmax distribution $H_\mathcal{G}(\mathcal{G}(S))$ is maximized, $\mathcal{L}_{pd} = [\sum_{S\in B}-H_{\mathcal{G}}(\mathcal{G}(S))]/|B|$.
$\mathcal{L}_{sp}$ is a segment smoothness penalty loss which encourages the generator to produce similar outputs for adjacent segments, $\mathcal{L}_{sp} = \sum_{(p_t, p_{t+1}) \in \mathcal{G}(S)}\parallel p_t - p_{t+1}\parallel^2$.
In the discriminator $\mathcal{C}$'s turn, the loss function to be \emph{minimized} is:
\begin{equation}
    \mathcal{L}_\mathcal{C}=\text{BCE}(\mathcal{C}(\mathcal{G}(S)), 1) + \text{BCE}(\mathcal{C}(P^r), 0) + \lambda\mathcal{L}_{gp}. \label{eq:loss_d}
\end{equation}
$\mathcal{L}_{gp}$ stands for gradient penalty \cite{wgan-training-nips17} that penalizes the gradient norm of the discriminator with respect to the input. An input sample $\tilde{P}\sim \tilde{\mathcal{P}}$ is a stochastic linear combination of the activations of pairs of real and fake samples that are cut down to the same shape, $\mathcal{L}_{gp}  = \mathbb{E}_{\tilde{P}\sim \tilde{\mathcal{P}}} \left[ ( \parallel \nabla_{\tilde{P}}\mathcal{C}(\tilde{P})  \parallel_2 - 1)^2\right]$ and $\tilde{P}  = \alpha P^r + (1-\alpha) \mathcal{G}(S)$.
Here, $\alpha \sim U(0, 1)$ stands for the stochastic weight. Hyperparameters $\eta, \gamma, \lambda$ are tuned by referring two unsupervised cross-validation metrics, LM negative log-likelihood and vocabulary usage \cite{wav2vecu_simp_NEURIPS2021_ea159dc9}.

\subsection{Diffusion GANs}
\label{subsec:diffgan}

In the wav2vec-U framework, the pseudo textual reference is sampled from a text set which does not necessarily align with the unlabeled audio set. Vanishing gradient occurs when the data and generator distributions are too different \cite{wang2022diffusiongan}. Again, we are suggested to tackle the training instability and mode collapse problems.  

We are interested in answering the following question: will diffusion processes \cite{ddpm_simp_NEURIPS2020_4c5bcfec} help improve the training stability and mode diversity for current adversarial training in wav2vec-U? To answer this, we seek to utilize an existing diffusion-GAN \cite{wang2022diffusiongan} framework, which includes three components, an adaptive diffusion process, a diffusion timestep-dependent discriminator, and a generator. Diffusion-GAN has been successfully applied in image generation and we adapt its formulation in a wav2vec-U scenario.

Concretely, both $P^r$ and $\mathcal{G}(S)$ are diffused by the same adaptive diffusion process. At each timestep $t$, the $t$-dependent discriminator $\mathcal{C}(\textbf{y}, t)$ learns to distinguish the \emph{diffused real data} from the \emph{diffused generated data}. $\mathcal{G}$ learns from each $\mathcal{C}(\textbf{y}, t)$'s feedback by back propagating through the forward diffusion chain with an adaptive length for noise-to-data ratio balancing. 

Diffusion-GAN alternately trains $\mathcal{G}$ and $\mathcal{C}(\cdot, t)$ by solving a min-max game objective:
\begin{equation}
    \min_\mathcal{G} \max_\mathcal{C} \mathbb{E}_{\textbf{x}, t, \textbf{y}}[\log \mathcal{C}_\varphi(\textbf{y}, t)] + \mathbb{E}_{\textbf{z}, t, \textbf{y}_\mathcal{G}}[\log (1-\mathcal{C}_\varphi(\textbf{y}_\mathcal{G}, t))]. \label{eq:min_max_diffgan}
\end{equation}
In the first item of Equation \ref{eq:min_max_diffgan}, $\textbf{x} \sim p(\textbf{x})$ and $p(\textbf{x})$ is the ``true'' data (i.e., the unlabeled $P^r$) distribution, $t \sim p_\pi$ and $p_\pi$ is a discrete distribution that assigns different weights $\pi_t$ to each diffusion timestep $t\in \{1, ..., T\}$, and $\textbf{y} \sim q(\textbf{y}|\textbf{x},t)$ where $q(\textbf{y}|\textbf{x},t)$ is the conditional distribution of the perturbed sample $\textbf{y}$ given the original data \textbf{x} (=$P^r$) and the diffusion timestep $t$. 
 
Recall that the forward diffusion process \cite{ddpm_simp_NEURIPS2020_4c5bcfec, DPM2015_DBLP:journals/corr/Sohl-DicksteinW15} gradually adds (Gaussian) noise to the ``true'' data $\textbf{x}=\textbf{x}_0 \sim p(\textbf{x}_0)$ in $T$ steps with a predefined variance schedule $\{ \beta_t \in (0, 1) \}_{t=1}^T$,  
    $q(\textbf{x}_{1:T} | \textbf{x}_0) := \prod_{t=1}^Tq(\textbf{x}_t | \textbf{x}_{t-1})$, and $q(\textbf{x}_t | \textbf{x}_{t-1}) = \mathcal{N}(\textbf{x}_t; \sqrt{1-\beta_t}\textbf{x}_{t-1}, \beta_t \textbf{I})$. 
With this setting, also by applying reparameterization trick and by merging two Gaussian noises, we can obtain $\textbf{x}_t=\textbf{y}$ at an arbitrary timestep $t$ as expressed by $q(\textbf{y}|\textbf{x}, t)$ in a closed form and by defining $\alpha_t:= 1-\beta_t$, $\bar{\alpha}_t:=\prod_{s=1}^t\alpha_s$. Thus, $q(\textbf{y}|\textbf{x}) :=\sum_{t=1}^T \pi_t q(\textbf{y}|\textbf{x},t)$, where 
    $q(\textbf{y}|\textbf{x},t) = \mathcal{N}(\textbf{y}; \sqrt{\bar{\alpha}_t}\textbf{x}, (1-\bar{\alpha}_t)\textbf{I})$.

In the second item of Equation \ref{eq:min_max_diffgan}, $\textbf{z} \sim p(\textbf{z})$ where $p(\textbf{z})$ is the unlabeled voice segment $S$'s distribution, and $\textbf{y}_\mathcal{G} \sim q(\textbf{y}|\mathcal{G}_\phi(\textbf{z}),t)$ where $q(\textbf{y}|\mathcal{G}_\phi(\textbf{z}),t)$ is the conditional distribution of the perturbed sample $\textbf{y}$ given the original generated data $\mathcal{G}_\phi(\textbf{z})$ (=$\mathcal{G}(S)$) and the diffusion timestep $t$.

Here, $\phi$ and $\varphi$ stand for the trainable parameter sets used in the generator $\mathcal{G}$ and the discriminator set $\mathcal{C}(\cdot, t)$, respectively.

\section{Diffusion-GAN Enhanced Wav2vec-U}
\label{sec:diffgan_wav2vecu}

\begin{figure*}
  \centering
  \includegraphics[width=17.5cm]{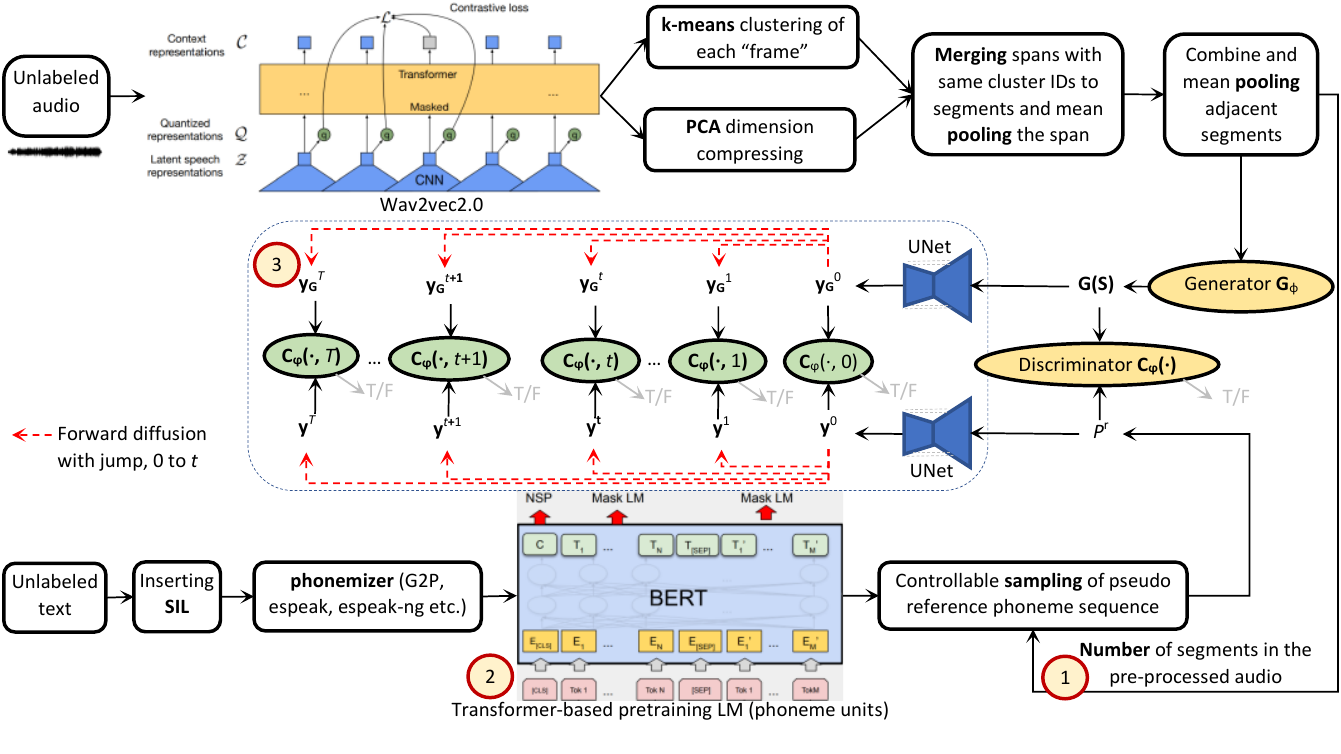}
  \caption{The training pipeline of our diffusion-GAN enhanced wav2vec-U system. Updates are marked in red-line cycled numbers.}
  \label{fig:diffgan_wav2vecU_framework}
\end{figure*}

The major idea of our proposed approach is to replace the vanilla GAN objectives used in wav2vec-U (Equation \ref{eq:loss_g} and \ref{eq:loss_d}) by the diffusion-GAN's objective (Equation \ref{eq:min_max_diffgan}). We depict the system architecture of our diffusion-GAN enhanced wav2vec-U pipeline in Figure \ref{fig:diffgan_wav2vecU_framework}.

There are three modules connected by two lines for (pre-)processing the unlabeled audio and the unlabeled text, respectively. The preprocessing of audios is the same as that used in wav2vec-U. The parameter weights in the pretrained wav2vec2.0 model are frozen during training the model. We pretrain a transformer-style phoneme-based language model in BERT \cite{bert-devlin-etal-2019-bert-simp} style to learn the phonemes' contextual distribution. From this distribution, we manage to perform controllable sampling \cite{bert-sampling-wang-cho-2019-bert} of pseudo phoneme sequence $P^r$ guided by the number of segments from the audio side. Note that $P^r$'s sampling can be performed offline before training. For each audio, we prepare a $\{P^r\}$ set that contains $a\times E$ elements where $a$ (=5) is a positive integer and $E$ is the number of training epochs. 

Since $\mathcal{G}(S)$ and $P^r$ are frequently expressed by relatively high dimension vectors (e.g., 512 in wav2vec-U baseline), we leverage a unsymmetrical U-Net \cite{Unet_DBLP:journals/corr/RonnebergerFB15} to ``down-sample'' (512, 256, 128 and 64) and ``up-sample'' (64 and 128) this high dimension gradually into a much lower dimension space to perform forward diffusion processes and determinations.

The generator $\mathcal{G}_\phi$ inherits from wav2vec-U and includes a single non-causal convolution layer. We also keep the original discriminator $\mathcal{C}_\varphi$ with three convolutional layers to distinguish the pseudo reference $P^r$ from the generator's prediction. The major updated part is the introducing of additional $T+1$ ($t\in [0, T]$) timestep-dependent discriminators $\mathcal{C}_\varphi(\cdot, t)$ that struggle at distinguishing the diffused ``real'' data $\textbf{y}^t$ from the diffused generated data $\textbf{y}_\mathcal{G}^t$. The generator learns from all these $T+2$ discriminators' feedback by back-propagating through the forward diffusion process. 

To better understand the theoretical properties of diffusion-GAN, two theorems were proved in \cite{wang2022diffusiongan}. The first theorem ensures that adding noise to the (pseudo) real and generated samples in a diffusion process can \emph{facilitate} the (adversarial) learning. The second theorem shows that minimizing the Jensen–Shannon (JS) divergence \cite{js-divergence-1207388} between the joint distributions of the noisy samples and the noise intensities, $p(\textbf{y}, t)$ and $p(\textbf{y}_\mathcal{G}, t)$, can lead to the same optimal generator as minimizing the JS-divergence between the original distributions of the (pseudo) real and generated samples, $p(P^r)$ and $p(\mathcal{G}(S))$. These two theorems hold in our architecture since our adaptation of diffusion-GAN from image generation domain to unsupervised ASR domain does not bring additional assumptions or pre-conditions. There are similar ideas of introducing multiple scales of noises perturbations simultaneously to train a single conditional score network for \emph{score} ($:= \nabla_{\textbf{x}}\log p(\textbf{x})$) estimating \cite{ScoreMatching_NEURIPS2019_3001ef25} of an unknown data distribution $p_{\text{data}}(\textbf{x})$.

As mentioned previously, instead of randomly sampling $P^r$ from a text set, we sample $P^r$ from a transformer-style \cite{transformer_NIPS2017_3f5ee243} phoneme-based pretrained language model $\mathcal{M}$. The sampling algorithms follow \cite{bert-sampling-wang-cho-2019-bert}. The motivation is that, when we pick $P^r$ from a text set, it can be arbitrarily long or short. Consequently, this sampling easily causes absurd length mismatching with the audio sequence which are given during training. To ease the unpredictable length mismatching, we first obtain the number of segments after pre-processing the audio using wav2vec2.0 representing + k-means clustering + segment merging, and then use this length to guide the length of sampled phoneme sequence from $\mathcal{M}$. This $P^r$ sampling method has several benefits during training: controllable pseudo reference text length, a better coverage of the target phoneme distribution, and a dense instead of the original one-hot representation of $P^r$ with richer contexts. 

\section{Experiments}

\subsection{Setup}

Our system largely leverages the github repositories of wav2vec-U\footnote{\url{https://github.com/facebookresearch/fairseq/blob/main/examples/wav2vec/unsupervised/README.md}}, Diffusion-GAN\footnote{\url{https://github.com/Zhendong-Wang/Diffusion-GAN}} and their default configurations of neural networks and hyperparameters. We take wav2vec-U as our major baseline and compare with it under three corpora, the TIMIT English dataset with five hours of audio recording annotated by time-aligned phonetic transcripts, the Librispeech dataset with 960 hours of labeled English speech audio, and the Multilingual LibriSpeech (MLS) dataset \cite{mls-dataset-Pratap_2020} that includes 6K hours for seven non-English languages. 

We let the discriminators learn from the less diffused sample pairs first and then gradually increase the difficulty by feeding them samples from larger $t$. Following the adaptive strategy with uniform $t\sim p_\pi$ in Diffusion-GAN, we count on a metric $r_d=\mathbb{E}_{\textbf{y},t}[\text{sign}(\mathcal{C}_\varphi(\textbf{y}, t)-0.5)]$ to estimate how much current $\mathcal{C}$ overfits to the data. Then, $T=T+\text{sign}(r_d-d_{\text{target}})*C$. We set $d_{\text{target}}=0.6$, $C=B*a_i/a_k$, $B$=64 for batch size, $a_i$=4 for updating interval and $a_k$=100, $\beta_0=0.0001, \beta_T=0.01, T_{\text{min}}=5, T_{\text{max}}=100$. All our experiments were performed under NVIDIA DGX-A100 with 8*A100-80GB GPUs.

\subsection{Librispeech Comparisons}

We first report results on the Librispeech benchmark to compare to competitive supervised, semi-supervised and wav2vec-U baselines. The language modeling dataset of Librispeech is selected as unlabeled text data for self-supervised training of a BERT-base \cite{bert-devlin-etal-2019-bert-simp} model and for unsupervised training of our diffusion-GAN enhanced wav2vec-U system. 

The unlabeled audio sequences are represented by the 15-th layer's output tensors of the 24-layer wav2vec2.0-Large model, which was trained on the 53.2K hours of Libri-Light (LL-60K) corpus. Other configures such as $\eta$, $\gamma$, $\lambda$, k (=128) in k-means and output dimension (=512) of PCA follows wav2vec-U.

Silence symbols \emph{SIL} are randomly inserted between two words with a probability of $p_{SIL}$ (=0.25) to align with the possible pauses and breaths in audio. We use the G2P phonemizer \cite{g2pE2019} which first annotates English words by the CMU pronouncing dictionary and then train a neural network to predict a word's phoneme sequence. 

\begin{table}
\centering
\begin{tabular}{l|cccc|c}
\hline
& \multicolumn{2}{c}{dev} & \multicolumn{2}{c|}{test}  & \\
\textbf{Model} & clean & other & clean & other & LM\\
\hline\hline
\multicolumn{6}{l}{\emph{Supervised}} \\
ContextNet & \textbf{1.9} & \textbf{3.9} & \textbf{1.9} & 4.1 & LSTM\\
Conformer & 2.1 & 4.3 & \textbf{1.9} & \textbf{3.9} & LSTM \\
CitriNet & - & - & 2.0 & 4.7 & Transf.\\ 
\hline
\multicolumn{6}{l}{\emph{Self/Semi-supervised}} \\
NST & 1.6 & 3.4 & 1.7 & 3.4 & LSTM\\
w2v2.0 & 1.6 & 3.0 & 1.8 & 3.3 & Transf.\\
w2v2.0+NST & \textbf{1.3} & \textbf{2.6} & \textbf{1.4} & \textbf{2.6} & LSTM\\
\hline
\multicolumn{6}{l}{\emph{Unsupervised}} \\
\textcircled{1}=w2v-U-L & 13.3 & 15.1 & 13.8 & 18.0 & 4-gram\\
\textcircled{2}=\textcircled{1}+ST & 3.4 & 6.0 & 3.8 & 6.5& 4-gram \\
           & 3.2 & 5.5 & 3.4 & 5.9 & Transf.\\
\cdashline{1-6}
\textcircled{1}+D-GAN & 11.6 & 13.5 & 12.1 & 16.4 & 4-gram\\
\textcircled{2}+D-GAN & 3.1 & 5.6 & 3.4 & 6.0 & 4-gram\\
           & \textbf{3.0} & \textbf{5.2} & \textbf{3.1} & \textbf{5.6} & Transf.\\ 
\hline
\end{tabular}
\caption{WERs (\%) on the Librispeech dev/test sets.}
\label{tab:librispeech_compare_result}
\end{table}

The WERs of nine baselines and our three model variants are listed in Table \ref{tab:librispeech_compare_result}. For each learning strategy, we select three strong baselines. For supervised learning, both ContextNet \cite{Han2020ContextNetIC} and CitriNet \cite{Majumdar2021CitrinetCT} are pure convolution based models enhanced with the squeeze-excitation strategy \cite{se-DBLP:journals/corr/abs-1709-01507}. In Conformer \cite{conformer_gulati20_interspeech}, two macaron-like feed-forward layers with half-step residual connections sandwich the multi-head self-attention \cite{transformer_NIPS2017_3f5ee243} and convolution modules. 

Noisy student training (NST) \cite{nst-Park_2020} augments Librispeech data by LL-60K in an iterative self-training method. W2v2.0 is the wav2vec 2.0 \cite{NEURIPS2020_92d1e1eb_wav2vec2_simp} self-supervised method followed by fine-tuning for ASR. Its large size (L) model is used for the baseline ``w2v-U-L'' and our diffusion-GAN (D-GAN) enhanced wav2vec-U models. ST stands for the subsequent self-training after GAN training.

Our model performs better than w2v-U-L under three configurations: with or without ST and 4-gram or transformer language models. The absolute improvements ranges from 0.2\%  to 1.7\%, and the relative improvements ranges from 5.1\% to 12.8\%. These results reflect that injecting instance noises of various intensities and appending diffusion timestep-dependent discriminators during adversarial training are effective.


\subsection{Unsupervised Comparison on TIMIT and MLS}

\begin{table}
\centering
\begin{tabular}{l|ccc}
\hline
& \multicolumn{2}{c}{core} & all  \\
\textbf{Model} & dev & test & test \\
\hline\hline
\multicolumn{4}{l}{\emph{matched setup}} \\
w2v-U & 17.0 & 17.8 & 16.6  \\
w2v-U + ST & 11.3 & 12.0 & 11.3   \\
\cdashline{1-4}
w2v-U + D-GAN & 15.8 & 16.4 & 15.6  \\
w2v-U + D-GAN + ST & \textbf{10.9} & \textbf{11.5} & \textbf{11.0}  \\
\hline
\multicolumn{4}{l}{\emph{unmatched setup}} \\
w2v-U & 21.3 & 22.3 & 24.4  \\
w2v-U + ST & 13.8 & 15.0 & 18.6  \\
\cdashline{1-4}
w2v-U + D-GAN & 20.0 & 21.2 & 22.9  \\
\ \ \ \ (1) -w/o BERT sampling & 20.3 & 21.5 & 23.3 \\
\ \ \ \ (2) -w/o length guiding & 20.6 & 21.6 & 23.5 \\
\ \ \ \ (3) -w/o ($T+1$) $\mathcal{C_\varphi}$ & 20.9 & 21.8 & 23.9 \\
w2v-U + D-GAN + ST & \textbf{13.4} & \textbf{14.5} & \textbf{18.1}  \\
\hline
\end{tabular}
\caption{PERs (\%) on TIMIT's dev/test sets.}
\label{tab:timit_uasr_compare_result}
\end{table}

Table \ref{tab:timit_uasr_compare_result} reports phoneme error rates (PER) under TIMIT's matched and unmatched training data setups. The same 4-gram LM is used for these four models. We only compare with w2v-U since it has achieved significantly better results than former GAN or HMM based unsupervised methods \cite{uasr_gan_hmm_Chen2019CompletelyUP}. Under both matched and unmatched setups, our diffusion-GAN enhanced w2v-U models outperforms w2v-U, with absolution improvements from 0.3\% to 1.4\% and relative improvements from 2.7\% to 7.9\%. 


We compare six languages' WERs under the MLS dataset. The baseline wav2vec-U+ST achieved WERs of 11.8\% (German), 21.4\% (Dutch), 14.7\% (French), 11.3\% (Spanish), 26.3\% (Italian) and 26.3\% (Portuguese), respectively. When we enhance this baseline by diffusion-GAN, our WERs are 11.1\%, 20.8\%, 14.5\%, 11.0\%, 25.7\%, and 25.8\%, respectively. On average, the WERs are reduced from 18.6\% to 18.2\%.

\subsection{Ablation Study}

We perform an ablation study by respectively turning off the BERT sampling, length guiding, and $T+1$ diffusion timestep-dependent discriminators. Table \ref{tab:timit_uasr_compare_result} reports the new PERs under the w2v-U+D-GAN configuration. Without BERT sampling \cite{bert-sampling-wang-cho-2019-bert}, we select pseudo texts randomly following the w2v-U baseline's unmatched setup. Then, we keep using BERT sampling yet with a random length without guidance from the audio side. Finally, we only use the original discriminator and skip the $T+1$ discriminators. The changes of PERs show that the discriminators contributed the most. With length guidance, we can obtain better selection criteria of the checkpoint selecting as well since the length matching information is now also involved in checkpoint training. 

Finally, we turn the U-Net off to perform diffusing and discriminating on the original latent space. The PERs were kept the same yet the training time cost for each batch increased about 12\%. Note that our three updates in Figure \ref{fig:diffgan_wav2vecU_framework} focus on the training process and the discriminator side only. The generator remains the same, thus the relatively slow multi-step backward reconstruction inferences in Denoising Diffusion Probabilistic Models (DDPMs) \cite{ddpm_simp_NEURIPS2020_4c5bcfec} or Noise Conditional Score Network (NCSN) \cite{ScoreMatching_NEURIPS2019_3001ef25} are circumvented. 


\section{Conclusion}

We enhance the adversarial training method for unsupervised ASR by a diffusion-GAN \cite{wang2022diffusiongan} that injects instance noises of various intensities to the real and fake samples, asks diffusion timestep-dependent discriminators to separate them, and back propagates the gradients to update the generator. The WER/PER comparisons with wav2vec-U \cite{wav2vecu_simp_NEURIPS2021_ea159dc9} under Librispeech, TIMIT and six languages in MLS, show that our enhancement strategies work effectively. Enriching the generator with diffusion processes is taken as our future work.

\bibliographystyle{IEEEbib}
\bibliography{strings,refs}

\end{document}